\documentclass[runningheads]{llncs}
\usepackage{graphicx}
\usepackage{amsmath,amssymb} 

\begin{document}
\pagestyle{headings}


\title{Location Augmentation for CNN} 

\author{{Zhenyi Wang} \and{Olga Veksler} }

\institute{
{University of Western Ontario\\
1151 Richmond St\\
London, ON\\
Canada}
\and
{University of Waterloo\\
 200 University Ave W\\
 Waterloo, ON \\
 Canada}
}

\maketitle

\begin{abstract}
CNNs have made a tremendous impact on the field of computer vision in the last several years.  The main component of any CNN architecture is the convolution operation,
which  is  translation invariant by design. However, location in itself can be an important cue. For example, a salient object is more likely to be closer to  the center of the image, the sky  in the top part of an image, etc. To include the location cue for feature learning, we propose to augment the color image, the usual input to CNNs, with one or more channels that carry location information.  We test two approaches  for adding location information. 
In the first approach,  we incorporate location  directly, by including the row and column indexes as two additional channels to the input image.  In the second approach, we 
add location less directly by adding distance transform from the  center pixel as an additional channel to the input image. 
We perform experiments with both direct and indirect ways to encode location.
We show the advantage of  augmenting the standard color input with location related channels on the tasks of salient object segmentation,  semantic segmentation, and scene parsing. 
\end{abstract}

\section{Introduction}
\label{sec:intro}

The advances in Convolutional Neural Networks (CNNs)~\cite{Fukushima1980,LeCun:1989,Krizhevsky:NIPS2012} lead to an extraordinary success in computer vision applications in the past several years. CNNs excel at learning reliable features for a multitude of vision problems, such 
as image classification~\cite{Krizhevsky:NIPS2012,overfeat:2014,VeryDeep:Zisserman},
object detection~\cite{Girshick:2014,Girshick:FastRCNN15,Residual:CVPR16},
semantic segmentation~\cite{farabet2013pami,FCN:Darrell,ChenPKMY14}, etc.

Traditional CNN architecture consists of three main building blocks: convolutional layers, pooling layers, and fully connected layers~\cite{Krizhevsky:NIPS2012}.
CNNs based on these components are highly successful in image classification tasks.
For semantic segmentation, the work in~\cite{FCN:Darrell} popularized omitting the fully connected layers. Unlike previous work, this approach enabled producing semantic segmentation at  pixel level, albeit at a coarse scale, with a single feed-forward  CNN run.

Thus the principle building block of most semantic segmentation CNNs 
is the operation of convolution, translation invariant by design.
The input to a semantic segmentation CNN is the three color channels of an image.  
Thus the convolutional deep features learned by CNN architectures without
fully connected layers are translation invariant. Therefore, when learning a useful feature, the location information is largely lost. However, location can be an important cue. For example, a salient object is more likely to be in the center of the image, the sky is  more likely to be in the top part of an image, etc.  

We propose to include location information in the feature learning process by augmenting the three color channels with one or more additional channels that carry location information,
see Fig.~\ref{fig:teaser}. First we experiment with adding location information directly.
In particular, we augment the input image with two additional channels, one for image row and one for image column indexes,  see Fig.~\ref{fig:teaser}(b).  
To be useful, location information does not have to be precise. Thus we also experiment with adding location information less precisely, namely through the distance transform from the image center, see Fig.~\ref{fig:teaser}(c). Here only the location relative to the image center is known, but that might be sufficient in order to obtain improvement over the color input only. The advantage is that only one additional input channel is needed, and fewer weights need to be learned.

Before the deep learning revolution, augmentation of pixel colors with pixel coordinates was used in 
the context of unsupervised machine learning for general purpose image segmentation~\cite{Felzenszwalb-graph-based}. With color only clustering based segmentation, the resulting segments are color consistent but can be spatially incoherent, depending on the image content. Adding pixel coordinates to data points to be clustered helps to  obtain both color and spatially consistent image segments.

The first work that uses coordinate augmentation that we are aware of  for computer vision applications is in~\cite{NIPS2017_7040}.  They augment input data with coordinates, but their goal is physics simulator, not semantic segmentaiton.   Simultaneously with our work, in~\cite{NIPS2018_coords}  they point out the defficiencies of CNNs that are based on RGB image data only, and propose coordinate augmentation for object recognition tasks.
Another  related work is in~\cite{XuPCYH16}. They are
interested in user-assisted object segmentation. The user provides positive (inside the object) and
negative (outside the object) seeds. They 
transform the user-provided seeds into
two Euclidean distance maps which are then concatenated
with the color image channels  and are used to train a CNN for object segmentation. 
In a subsequent work~\cite{DeepGrabCut:XuPCYH17}, they extend similar ideas to the scenario of the user provided  loose (can be too small or too large) box around the object of interest. Euclidean distance map is constructed from the softened user box and added to the color channels of the image as an input to CNN.
The works in~\cite{XuPCYH16,DeepGrabCut:XuPCYH17} are similar in spirit but of a more narrow scope compared to our work.  Similar to our work,   in~\cite{XuPCYH16,DeepGrabCut:XuPCYH17} they are interested in adding a location channel to help to train a deep neural network. Unlike our work,  their location information comes from the user input only, whereas we are using general location information already contained in the data without any user interaction. 
In computational biology, ~\cite{Delong:NIPS:workshop} use distance transform idea to encode a distance  from certain features as an additional input track to CNN.

We evaluate our approach on the tasks of salient object segmentation, semantic segmentation, and scene parsing.  We show that both direct and indirect  location augmentation improve the accuracy compared to using RGB channels only. 
In most cases, adding the distance transform channel works better than adding two coordinate channels. 
The extra computational time needed when augmenting RGB input with location
information is small, especially  for the  networks that are more deep, since
location augmentation effects only the input layer.  
We also experimentally investigate the relationship between the depth of the network and the improved accuracy of adding location augmentation. 

This paper is organized as follows. In Sec.~\ref{sec:location} we motivate location augmentation.  In Sec.~\ref{sec:saliency}, \ref{sec:semantic}, \ref{sec:scene}
we show the advantage of including location information on the tasks of saliency segmentation,  semantic image segmentation and scene parsing. 
Conclusions are in Sec.~\ref{sec:conclusions}.

\section{Location Augmentation}
\label{sec:location}

\begin{figure}
\begin{tabular}{ccc}
\fbox{\includegraphics[height=1.7cm]{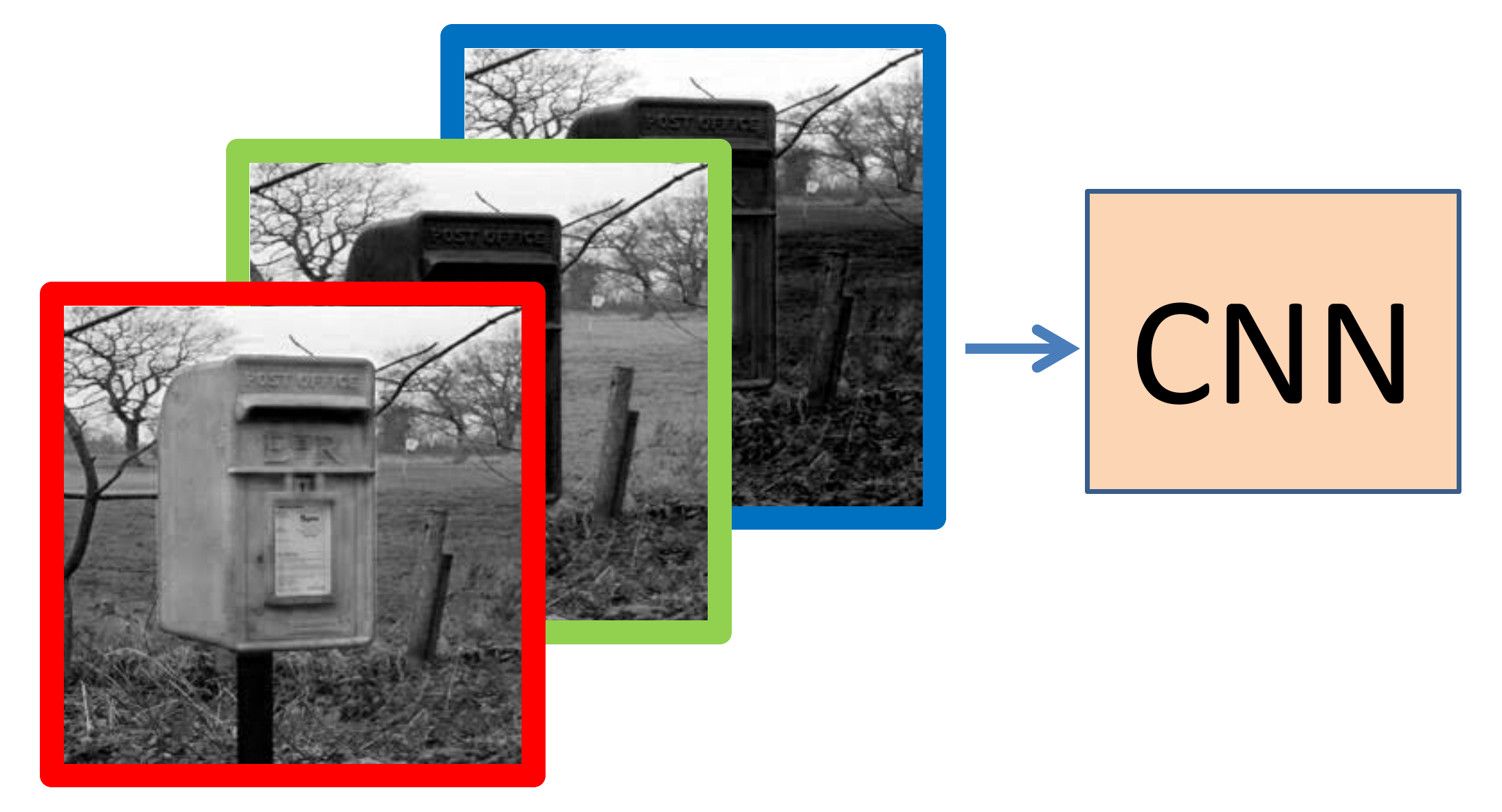}}&
\fbox{\includegraphics[height=1.7cm]{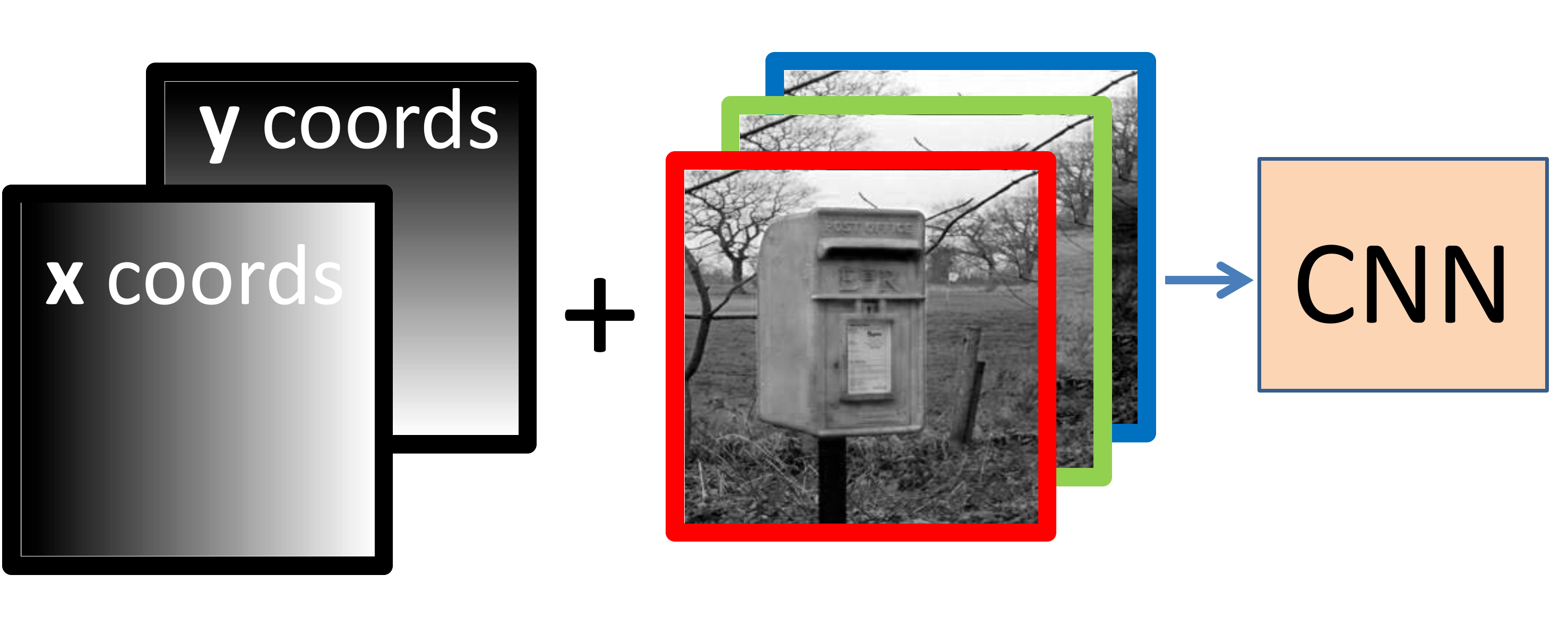}}&
\fbox{\includegraphics[height=1.7cm]{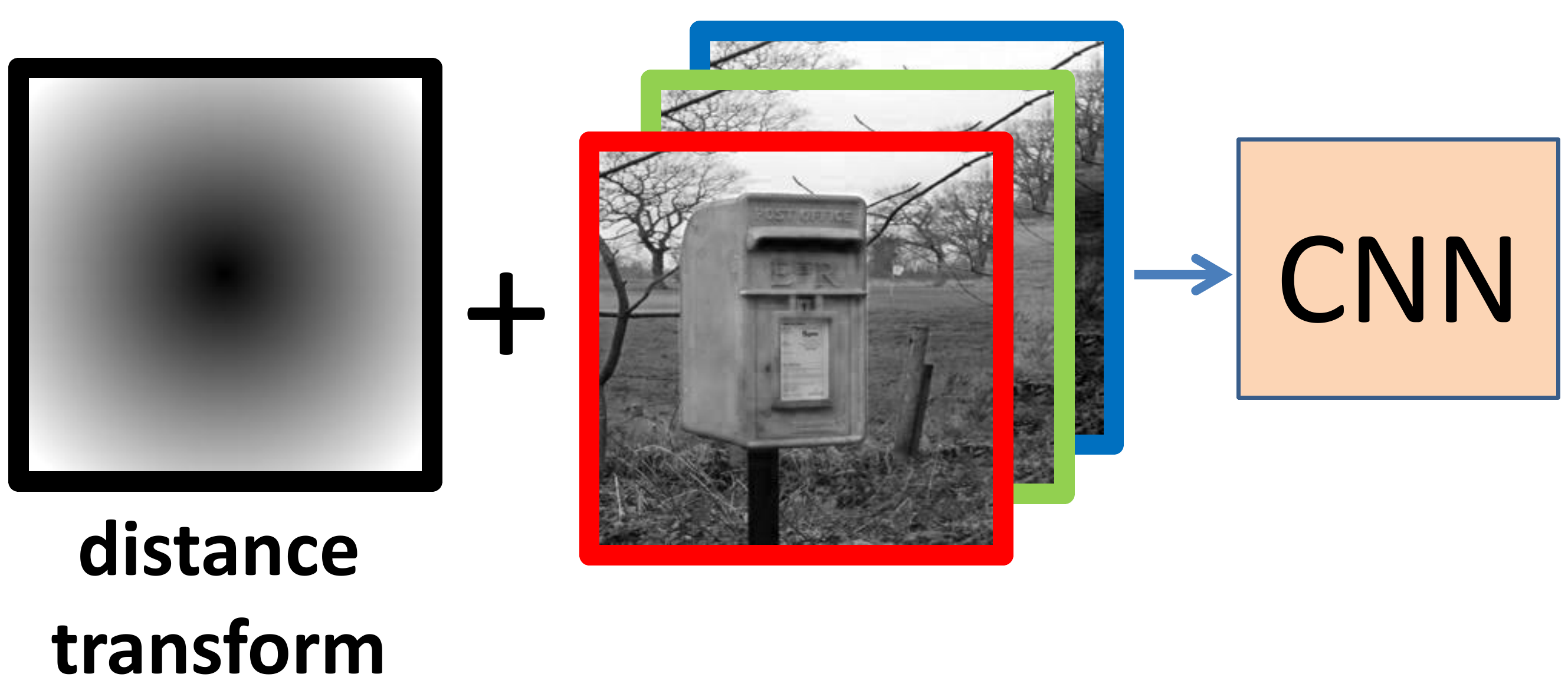}}\\
(a)&(b)&(c)
\end{tabular}
\caption{Usually the input to a fully convolutional CNN is the three channels of a color image, shown in (a). We propose to augment the input with location information.  In (b) we directly
  augment the image with the $x$ and $y$ image coordinates as two separate input channels.
In (c) we add location information less directly, by adding the distance transform from the central pixel as an additional channel.  }
\label{fig:teaser}
\end{figure}

For location usefulness motivation, consider the following artificial machine learning
problem.
Suppose we have a semantic segmentation task with just two classes: the foreground and the background. The  training examples consist of a random collection of distinct color images, and the ground truth is some random shape, for example a circle, of exactly the same radius and in exactly the same location for all the images in the dataset. Assume CNN consists only of convolutional layers  with filters of small size, say  3 by 3.
With the location information included, learning to segment the circle shape is a trivial task even for for a shallow CNN. The correct solution can be learned with just one hidden layer and $3$ by $3$ convolution
filters.

 Without the location information, learning to segment the circle shape is difficult with a CNN, because whether a pixel belong to the circle object depends only on its location, not the color.  Interestingly, with a deep CNN and zero padding in the convolutional layers, the circle shape can be learned to some degree, depending
on the distribution of colors in the dataset. Zero padding gives a location cue for pixels at image border. 
This location cue can be eventually propagated to the rest of the pixels if the network is deep enough.  However, for such a trivial learning task, having to employ a deep network is not an intellectually satisfying solution, as well as a computationally expensive one.  A cheap shallow network  on location augmented images is a more sensible solution.

We propose to include location information in the feature learning process by augmenting the three color channels with one or more additional channels that carry location information,
see Fig.~\ref{fig:teaser}.  
In Fig.~\ref{fig:teaser}(b), the row and column indexes are added directly as separate channels. 
Instead of using two separate channels, we also tried 
to encode the precise location information in just one channel, by linearly indexing all image pixels and providing the linear index in one additional channel. However, we found that CNN with such an augmentation is not easy to train in practice. Since the linear index has a much larger scale than image color,
we normalized all input features (color and pixel index) to be in the same range, but that did not help training.  The reason training is more difficult, therefore, is likely to be that the useful information about pixel position, for example is whether a pixel is on the left or on the right of the image involves division and modulo arithmetic and might be difficult to learn from linear indexes. In contrast, such location information is easy to learn from the separate channels for row and column indexes as in  Fig.~\ref{fig:teaser}(b).

In Fig.~\ref{fig:teaser}(c) we illustrate augmenting image colors with an additional channel that carries only approximate location information. In particular, it is the distance transform from the image center. This allows learning features that are based on proximity to image border and/or image center. 
We can also combine the row/column channels in Fig.~\ref{fig:teaser}(b) with the distance transform
channel Fig.~\ref{fig:teaser}(c) to have three additional input channels.

\section{Experimental Evaluation}
In this section, we evaluate location augmentation on the applications of saliency, semantic segmentation and scene parsing. We describe implementation details of our method, introduce datasets and evaluation criteria, and compare the performance on different tasks and datasets by our proposed approach.

In the following sections, all experiments are done on a computer with 16GB main memory and GeForce GTX 1080 with 8GB GPU memory. We use the original datasets without any augmentation with additional data.

\subsection {Salient Object Segmentation }
\label{sec:saliency}

\begin{table}[h]\centering
\begin{tabular}{|l|c|c|c|}
\hline
&HKU-IS\cite{guanbin15}&ECSSD\cite{qiong13}&PASCAL-S\cite{yinli14}\\
\hline
RGB &0.5314  & 0.5006& 0.3854\\
\hline
RGB+dist &0.5908  & 0.6199& 0.4831\\
\hline
RGB+coord &0.5813  & 0.5764& 0.4403\\
\hline
RGB+dist+coord &\bf{0.6081}  & \bf{0.6358}& \bf{0.4948}\\
\hline
\end{tabular}
\caption{F-measure of CNN with  2 pooling layers.}\label{tab:saliency1}
\end{table}

\begin{table}[!hbp]\centering
\begin{tabular}{|c|c|c|c|}
\hline
 &HKU-IS\cite{guanbin15}&ECSSD\cite{qiong13}&PASCAL-S\cite{yinli14}\\
\hline
RGB &0.6653  &0.6786 &0.5164 \\
\hline
RGB+dist &\bf{0.7059}  &\bf{ 0.7202} &\bf{0.5458} \\
\hline
RGB+coord &0.6873  &0.7031 &0.5299 \\
\hline
RGB+dist+coord &0.6965  &0.7131 &0.5399 \\
\hline
\end{tabular}
\caption{F-measure of CNN with  3 pooling layers.}\label{tab:saliency2}
\end{table}

\begin{table}[!hbp]\centering
\begin{tabular}{|c|c|c|c|}
\hline
&HKU-IS\cite{guanbin15}&ECSSD\cite{qiong13}&PASCAL-S\cite{yinli14}\\
\hline
RGB &0.6832  &0.7018 &0.5261 \\
\hline
RGB+dist &\bf{0.7228}  &\bf{0.7445} &\bf{0.5611} \\
\hline
RGB+coord &0.7213  &0.7342 &0.5507 \\
\hline
RGB+dist+coord &0.7155  &0.7344 &0.5528 \\
\hline
\end{tabular}
\caption{F-measure of CNN with 4 pooling layers.}\label{tab:saliency3}
\end{table}

\begin{table}[!hbp]\centering
\begin{tabular}{|c|c|c|c|}
\hline
&HKU-IS\cite{guanbin15}&ECSSD\cite{qiong13}&PASCAL-S\cite{yinli14}\\
\hline
RGB &0.7061  &0.7240 &0.5406 \\
\hline
RGB+dist &\bf{0.7266}  &\bf{0.7493} &\bf{0.5619} \\
\hline
RGB+coord &0.7235&0.7368& 0.5517\\
\hline
RGB+dist+coord &0.7167  &0.7352 & 0.5541\\
\hline
\end{tabular}
\caption{F-measure of CNN with 5 pooling layers.}\label{tab:saliency4}
\end{table}

\textbf{Datasets}. We evaluate our method on several representative datasets, including MSRA-B~\cite{tie11}, ECSSD\cite{qiong13}, HKU-IS ~\cite{guanbin15}, PASCAL-S~\cite{yinli14}, all of which are available online. These datasets contain many images with diverse scenes and have been widely used as saliency segmentation benchmark. MSRA-B contains 5000 images from hundreds of categories. Most images in this dataset contain only one salient object. Due to its diversity and large quantity, MSRA-B has been one of the most widely used datasets in salient object segmentation literature. ECSSD contains 1000 semantically meaningful but structurally complex natural images. HKU-IS is another large-scale dataset that contains more than 4000 challenging images. Most images in this dataset have low contrast with more than one salient object. PASCAL-S contains 850 challenging images (each composed of several objects), all of which are chosen from the validation set of the PASCAL VOC 2010 segmentation dataset. All these datasets provide ground truth human annotations.
All  networks in this section are trained on the entire MSRA-B dataset, i.e., 5000 images,
and tested on ECSSD, HKU-IS and PASCAL-S datasets.

\begin{figure}
\begin{tabular}{cccc}
\fbox{\includegraphics[width=2.3cm]{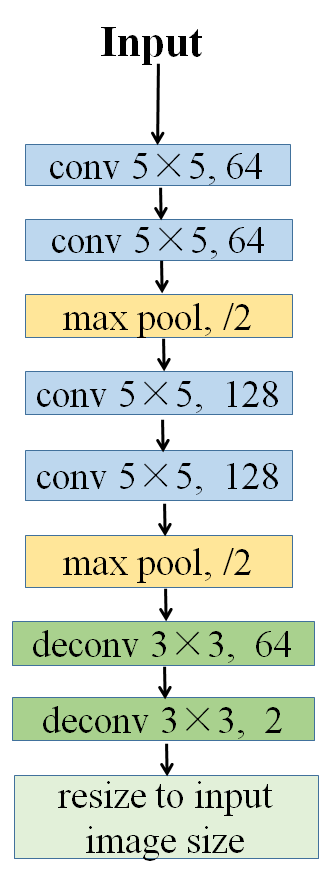}}&
\fbox{\includegraphics[width=2.3cm]{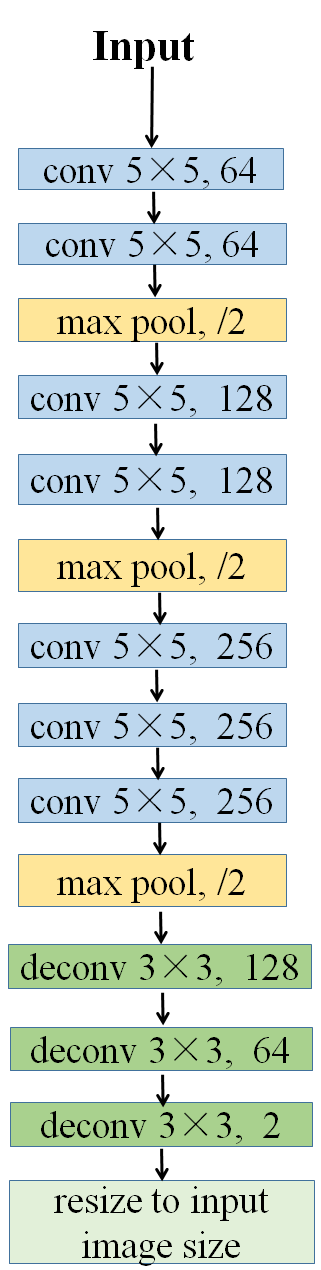}}&
\fbox{\includegraphics[width=2.3cm]{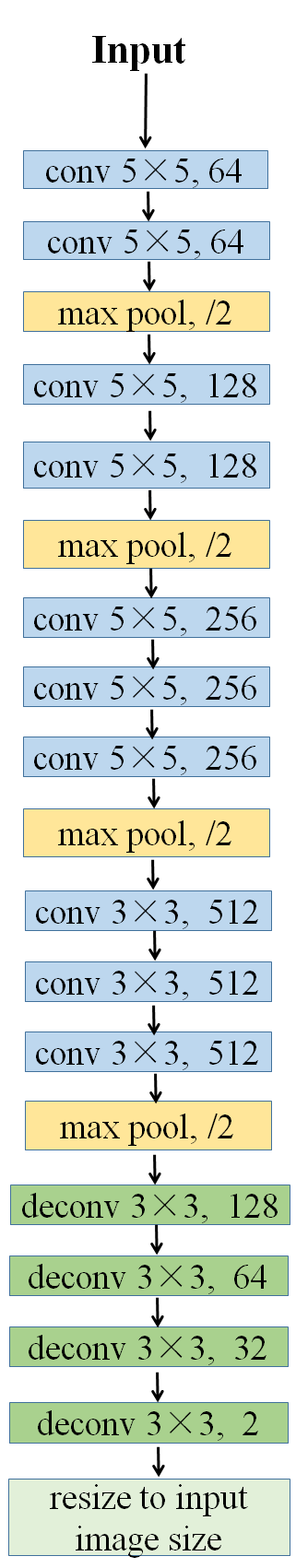}}&
\fbox{\includegraphics[width=2.3cm]{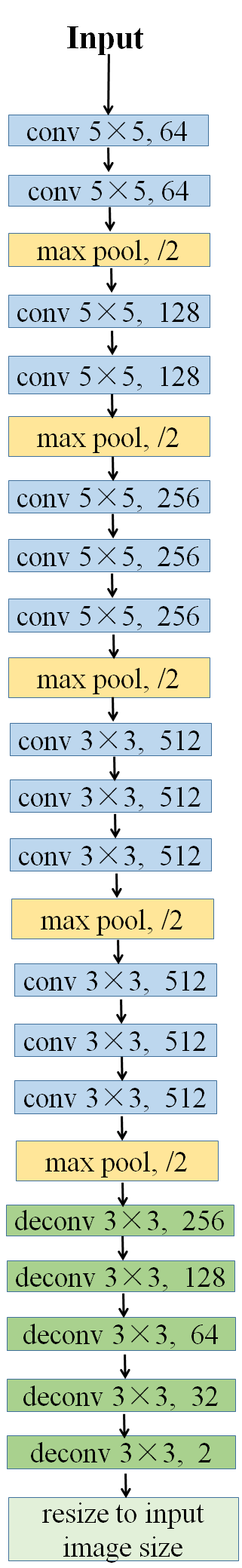}}\\
(a)&(b)&(c)&(d)
\end{tabular}
\caption{Saliency segmentation networks with different number of pooling layers: (a)  2 pooling layers; (b) 3 pooling layers; (c)  4 pooling layers; (d)  5 pooling layers.}
\label{fig:network}
\end{figure}

\textbf{CNN Architecture}.
We test the effectiveness of location augmentation by experimenting with networks of
different depth. We use the standard decoder-encoder architecture~\cite{FCN:Darrell}
with 2, 3, 4, and 5 pooling layers, illustrated in Fig.~\ref{fig:network}.

For each network,  we test the standard RGB input (Fig.~\ref{fig:teaser}(a)), 
RGB augmented with coordinates (Fig.~\ref{fig:teaser}(b)), RGB augmented
with distance transform from the center (Fig.~\ref{fig:teaser}(c)), and, finally,
RGB augmented with both, the coordinates and the distance transform. In all the 
figures and tables, we refer to these different input types as `RGB', `RGB+coord', `RGB+dist', `RGB+dist+coord', respectively.

\textbf{Implementation Details.} Our networks are implemented with TensorFlow~\cite{tensorflow2015-whitepaper}. The mini-batch size is set to be 2 and learning rate is set to be $10^{-4}$  and weight decay is set to $10^{-6}$. We use Adam~\cite{Kingma15} to train the network from scratch without utilizing pretrained model weights.

\textbf{Performance Metrics}. To evaluate the quality of a saliency map, we use the F-measure, which is defined as

\begin{equation}
F_{\beta}=\frac{(1+\beta^{2})Precision \times Recall}{\beta^{2}Precision+Recall}
\end{equation}

As published in previous research articles on saliency segmentation, we also set $\beta^{2}$ to be 0.3 for stressing the importance of the precision value.

\textbf{Inference Time}. The running time  is evaluated by  averaging
over 1000 images with each size of 400 $\times$ 300 and 10 trials for each input on a GeForce GTX 1080. The running time is calculated in  seconds.

\textbf{Discussion} Tables~\ref{tab:saliency1}, \ref{tab:saliency2}, \ref{tab:saliency3} and~\ref{tab:saliency4} show the F-measure obtained by training the networks with 2, 3, 4,   and 5 pooling layers, respectively.  In all cases, augmenting RGB channels with location information helps. The biggest F-measure gains are on the network with 2 pooling layers, from around 7\% to 13\%, depending on the dataset,  see Table~\ref{tab:saliency1}. But for deeper networks, with more pooling layers,  F-measure gains remain significant, around 2\% in F-measure across different dataset for the deepest network with 5 pooling layers. 
Comparing different location augmentation methods,  
 for the network with 2 pooling layers,
RGB+dist+coords is the best performer, and for networks with 3, 4, and 5  
pooling layers, 
RGB+dist is the best performer. 
Running times on all the networks are in Table~\ref{tab:runtime}. The increase in
inference time for location augmented input is small, especially as the networks get deeper. 
Thus location augmentation can be used to increase accuracy, or to use a less deep network without losing accuracy but gaining computational efficiency.

\begin{table}[!hbp]\centering
\begin{tabular}{|c|c|c|c|c|}
\hline
&2 pooling &3 pooling &4 pooling &5 pooling\\
\hline
RGB &0.0173  &0.0284 &0.0308& 0.0350\\
\hline
RGB+dist &0.0175  &0.0287 &0.0312 & 0.0352\\
\hline
RGB+coord &0.0182  &0.0292 &0.0318 & 0.0356\\
\hline
RGB+dist+coord &0.0192  &0.0298 &0.0327& 0.0361\\
\hline
\end{tabular}
\caption{Running time on one image on networks with different number of pooling layers and input channels.}\label{tab:runtime}
\end{table}

\subsection {Semantic Segmentation}
\label{sec:semantic}

In semantic segmentation, the task is to assign each image pixel an object class.
Compared to saliency segmentation, it is a  more challenging task due to a larger number of classes. We use the mean of class-wise intersection over union (Mean IoU), a standard metric in semantic segmentation, for evaluation.

\begin{table}[!hbp]\centering
\begin{tabular}{|c|c|c|c|c|c|c|c|c|c|c|c|c|c|}
\hline
&aeroplane& bicycle&bird& boat& bottle&bus&car\\
\hline
RGB &69.1&  33.0&  72.8&  53.9& 70.7&  \bf{76.8}&  72.1 \\
\hline
RGB+dist &75.7&  \bf{34.1}&  75.8&  \bf{59.0}& 71.7&  75.8&  \bf{76.2} \\
\hline
RGB+coord &75.7&  32.9&  76.2&  55.1& 70.1&  75.2&  73.9\\
\hline
RGB+dist+coord &\bf{77.6}&  33.8&  \bf{77.7}&  54.9& \bf{72.7}&  75.1&  74.5\\
\hline
\end{tabular}
\caption{Per-class IoU on  a subset of PASCAL VOC 2011 segval.}\label{tab:perclass1}
\end{table}

\begin{table}[!hbp]\centering
\begin{tabular}{|c|c|c|c|c|c|c|c|c|c|c|c|c|c|}
\hline
&cat &chair&cow&diningtable&dog& horse&motorbike\\
\hline
RGB&75.0 &24.5 &62.0&  35.0&  66.1&  58.8&  75.6\\
\hline
RGB+dist  &78.1 &24.5 &\bf{69.2}&  36.6&  69.8&  \bf{65.2}&  \bf{77.9}\\
\hline
RGB+coord  &77.0 &\bf{26.6} &64.0&  35.9&  67.5&  61.6&  75.7\\
\hline
RGB+dist+coord  &\bf{78.9} &23.8 &64.4&  \bf{36.8}&  \bf{70.0}&  63.4&  75.2\\
\hline
\end{tabular}
\caption{Per-class IoU on  a subset of PASCAL VOC 2011 segval.}\label{tab:perclass2}
\end{table}

\begin{table}[!hbp]\centering
\begin{tabular}{|c|c|c|c|c|c|c|c|c|c|c|c|c|c|}
\hline
& person&potted plant & sheep &sofa &train &tv&mean IoU\\
\hline
RGB &74.6 &41.9 &68.3&  32.1&  76.8 &55.6& 61.2\\
\hline
RGB+dist  &\bf{76.6} &\bf{46.6} &\bf{72.7}&  32.6&  76.7 &55.5& \bf{63.8}\\
\hline
RGB+coord   &75.8 &42.6 &69.5&  32.7&  75.5 &57.5& 62.4\\
\hline
RGB+dist+coord   &76.1 &42.6 &69.8&  \bf{35.1}&  \bf{76.9} &\bf{59.3}& 63.3\\
\hline
\end{tabular}
\caption{Per-class IoU on  a subset of PASCAL VOC 2011 segval.}\label{tab:perclass3}
\end{table}

We evaluate our proposed method on the PASCAL VOC dataset. The PASCAL VOC 2011 segmentation challenge contains 1112 training images. Hariharan et al.~\cite{Bharath11} collected labels for a larger set of 11318 PASCAL images. The images in the SBD are divided into 8498 training images and 2820 test images. The test images are from a subset of the VOC2011 validation set. There are training images from \cite{Bharath11} included in the PASCAL VOC 2011 validation set, so we test the networks on the non-intersecting set of 736 images as done in \cite{jonathan15}. We use FCN-32s ~\cite{jonathan15} as the base network and use Caffe \cite{yangqing14} to implement the network. We report the best results achieved after convergence at a fixed learning rate $1^{-10}$ by training using SGD with  momentum set to 0.99. Weight decay is set to 0.0005. Since pretrained weights are not available for location augmentation connections, to make a fair comparison, we use  randomly initialized the parameters for the input layer (whether the input is
RGB, or RGB+dist, etc.). 

The semantic segmentation results on SBD dataset are shown in 
Tables~\ref{tab:perclass1}, ~\ref{tab:perclass2}, and~\ref{tab:perclass3}.
The mean IoU is in the last column of Table~\ref{tab:perclass3}.
Here, again, all location augmentation CNNs perform better than CNNs with RGB input only. The best overall performer is RGB+dist, with the mean IOU 2.6\% better compared
to using RGB input. 

Looking at individual object classes, the airplane class improves the most, namely 
8.5\% with `dist+coord' augmentation over just RGB input. This makes sense, as airplanes tend to be in the top part of the image and location augmentation can help significantly. The bird class also shows a large improvement, almost 5\%, for the same reason. 
Other classes that show a large improvement with location added are cow, boat, and the horse classes. The only class that does not show any improvement  with location augmentation is the bus class. Interestingly, the car class does show a significant improvement with location, even though it is conceptually similar to the bus class. This happens probably because cars are distributed in less random locations in this dataset compared to the bus locations.

\subsection {Scene Parsing}
\label{sec:scene}

\begin{table}[!hbp]\centering
\begin{tabular}{|c|c|c|c|c|c|c|c|c|}
\hline
&road& sidewalk& building& wall& fence& pole& traffic light\\
\hline
RGB &94.7 &64.8& 82.7& 35.7&  38.0& 14.2& 19.3\\
\hline
RGB+dist &\bf{95.4} &\bf{69.1}&\bf{84.0}& 39.6& \bf{40.7}& \bf{17.8}& \bf{25.8}\\
\hline
RGB+coord &95.2  &67.3  &83.2  &37.7&  39.2& 16.9&  23.8\\
\hline
RGB+dist+coord &95.3  &68.2&  83.7&  \bf{40.4} &39.4 &17.2&  25.3\\
\hline
\end{tabular}
\caption{Per-class IoU on Cityscapes validation set.}\label{tab:city1}
\end{table}

\begin{table}[!hbp]\centering
\begin{tabular}{|c|c|c|c|c|c|c|c|c|}
\hline
&traffic sign &vegetation& terrain& sky& person& rider& car\\
\hline
RGB & 31.6 &82.8& 47.5& 82.2&  49.9&  26.2&  82.4\\
\hline
RGB+dist  & \bf{38.7} &\bf{84.2}& \bf{49.6}& 85.0&  \bf{52.9}&  29.3&  \bf{84.7}\\
\hline
RGB+coord &34.8& 83.4&  46.1& 84.9&  51.9&  29.5&  83.6\\
\hline
RGB+dist+coord &36.3&  83.8&  48.7& \bf{85.4}&  51.9& \bf{30.7}&  84.1\\
\hline
\end{tabular}
\caption{Per-class IoU on Cityscapes validation set.}\label{tab:city2}
\end{table}

\begin{table}[!hbp]\centering
\begin{tabular}{|c|c|c|c|c|c|c|c|c|}
\hline
& truck & bus &train&motorcycle&bicycle &mean IoU\\
\hline
RGB & 44.6 & 57.4& 38.7& 26.5&  47.5&50.9\\
\hline
RGB+dist & \bf{47.2} & \bf{59.2}& \bf{38.9}& \bf{34.6}&  \bf{52.8}&\bf{54.2}\\
\hline
RGB+coord &41.1 &51.2&  37.3&  31.1& 52.0&52.1\\
\hline
RGB+dist+coord &46.7& 56.1& 37.7  &32.4&  52.4 &53.5\\
\hline
\end{tabular}
\caption{Per-class IoU on Cityscapes validation set.}\label{tab:city3}
\end{table}
Scene parsing is  similar to semantic segmentation.  The goal is to assign each pixel in the image a category label. Scene parsing provides a complete understanding of the scene
and is useful  for automatic driving, robot sensing, etc.

We evaluate the performance of our proposed method on scene parsing task on
Cityscapes  dataset ~\cite{marius16}, which is a recently released large-scale dataset that contains a diverse set of street scenes for semantic urban scene understanding. It contains high quality pixel-level finely annotated images collected from 50 cities. We train the network on 2,975 training images, and test the network on its validation set which contains 500 images. These images define 19 categories containing both stuff and objects. All images are resized from 1024 $\times$ 2048 to 512 $\times$ 1024 to fit into computer GPU memory.

We also use FCN-32s ~\cite{jonathan15} as the base network and use Caffe \cite{yangqing14} to implement the network. We report the best results achieved after convergence at a fixed learning rate $2\cdot 10^{-12}$ by training using SGD with a high momentum of 0.99.  For a fair comparison, we also randomly initialize the parameters of first layer, since pretrained weights on location connections are not available.

The scene parsing results on Cityscapes dataset are shown in 
Tables~\ref{tab:city1}, \ref{tab:city2}, and \ref{tab:city3}. The mean IoU measure
over all classes is in the last column of Table~\ref{tab:city3}.
Again as in semantic segmentation, the best result is obtained with distance augmentation,
that is using RGB+dist for the input layer. 
The mean IoU measure is improved by 3.3\%. 
All classes show improvement using location augmentation. The most improved classes
are traffic sign, wall, sidewalk, motorcycle. The least improved class is the road.

\section{Conclusions}
\label{sec:conclusions}
In this paper, we proposed to include location information in the input layer of a CNN architecture, adding it to the standard RGB input.  We include a direct measure of location, such as image coordinates, and also an indirect measure, such as distance transform from the image center. We also test the combination of the two. In most cases, adding just the distance transform gives a better accuracy, compared to adding direct location information.
 This might be due to  the necessary  information about location being sufficiently represented by the distance from the image center, on the one hand, and the distance transform augmentation having less parameters to learn, on the other hand. 

We showed that a significant achievement in the accuracy can be achieved by adding 
location information, while the increase in the computational time is negligible.
The increase in computational time is small because the number of new weights introduced is small. 
The deeper is the network, the less effect location augmentation has, as expected, but still it stays significant as the depth is increased. 

\bibliographystyle{splncs}
\bibliography{accv2018}

\end{document}